# Introducing BEREL: BERT Embeddings for Rabbinic-Encoded Language

Avi Shmidman,[1,2] Joshua Guedalia,[1] Shaltiel Shmidman,[1]
Cheyn Shmuel Shmidman,[1] Eli Handel,[1] Moshe Koppel[1,2]

[1]DICTA: Israel Center for Text Analysis, Jerusalem, Israel
[2]Bar-Ilan University, Ramat-Gan, Israel
{avi.shmidman}@biu.ac.il
{joshua.guedalia,shaltiel.shmidman,cheyn.shmuel,eli.handel,moishk}@gmail.com

## Abstract

We present a new pre-trained language model (PLM) for Rabbinic Hebrew, termed *Berel* (BERT Embeddings for Rabbinic-Encoded Language). Whilst other PLMs exist for processing Hebrew texts (e.g., HeBERT, AlephBert), they are all trained on modern Hebrew texts, which diverges substantially from Rabbinic Hebrew in terms of its lexicographical, morphological, syntactic and orthographic norms. We demonstrate the superiority of *Berel* on Rabbinic texts via a challenge set of Hebrew homographs. We release the new model and homograph challenge set for unrestricted use.

## 1 Introduction

High-resource languages such as English enjoy many available pretrained language models (BERT (Devlin et al., 2019), RoBERTa (Liu et al., 2019), BART (Lewis et al., 2020), T5 (Raffel et al., 2020), and more. The past few years have seen the release of such models for Hebrew as well. The first such model was heBERT (Chriqui and Yahav (2021)). Afterward, AlephBERT (Seker et al., 2021) established the current SOTA for Hebrew pretrained contextualized embeddings. These models have shown great success on modern Hebrew texts, but when applied to Rabbinic Hebrew they fall short. In this paper, we introduce a new pretrained contextualized language model specifically trained for Rabbinic Hebrew.

## 2 The Challenge of Rabbinic Hebrew

Rabbinic Hebrew differs from modern Hebrew in several ways. First of all, Rabbinic Hebrew writers freely interweave Talmudic Aramaic within their texts. This includes not only integration of Aramaic phrases and passages, but also morphological conflation, wherein Aramaic prefixes and suffixes are attached productively to Hebrew words. Additionally, the orthography of Rabbinic Hebrew is far less predictable than modern Hebrew. While modern Hebrew adheres to a fairly standard plene orthography, Rabbinic Hebrew admits to a wide range of orthographic options, both plene and defective. As scholars have noted, modern Hebrew by itself is already highly ambiguous morphologically (Wintner, 2014; Tsarfaty et al., 2019); however, without the norms of plene spelling, the ambiguity is amplified considerably. Analogously, the morphological and syntactic norms of Rabbinic Hebrew are also far less standardized, and often diverge wildly from those of modern Hebrew. Due to all of the foregoing, a modern Hebrew language model will falter when presented with a Rabbinic Hebrew text.

An additional complication of Rabbinic Hebrew is the rampant use of abbreviated words, including both apocopated words (for example, אמרינן will often be written as אמרינ׳ or אמרי׳), as well as acronyms for multi-word sequences (for example, עד כאן לשונו will often be written as עכ"ל). In a typical printed Rabbinic text, one out of every 5-10 words is comprised of an abbreviation. However, the way such abbreviations are handled by the standard BERT word piece tokenizer is far from ideal: every apostrophe or quotation mark is considered a new token, and thus these frequent abbreviations will always be broken up into multiple pieces. This, too, hampers the ability of existing Hebrew BERT models when processing Rabbinic Hebrew.

## 3 A New Model: *Berel*

The new model we deliver is dubbed *Berel*: BERT Embeddings for Rabbinic Encoded Language. *Berel* is trained on the entirety of two large public-domain databases of Rabbinic texts: The Sefaria Library[1] and the Dicta Library[2]. Together, these databases contain approximately 220 million words of text (200 million in Sefaria and 20 million in the Dicta Library). We trained *Berel* with the BERT training script (Devlin et al., 2019), using the **BERT-base** configuration, and a vocabulary of 128,000 word pieces. Furthermore, we adjusted the word piece tokenization script such that, upon encountering an apostrophe or double quote, it first determines whether the mark is part of a Rabbinic Hebrew abbreviation, and if so, the tokenizer keeps

---

[1] https://www.sefaria.org/texts
[2] https://library.dicta.org.il/

| Word | Option 1 | Option 2 | W2V | AlephBERT | BEREL |
|---|---|---|---|---|---|
| אחר | אַחַר | אַחֵר | 79.40% | 93.76% | **94.09%** |
| בנה | בָּנָה | בָּנֶה | 82.36% | 97.00% | **100.00%** |
| הפר | הֵפֵר | הָפַר | 93.39% | 98.56% | **99.64%** |
| הפרה | הַפָּרָה | הֲפָרָה | 89.52% | 94.60% | **97.74%** |
| חדשים | חֳדָשִׁים | חֲדָשִׁים | 87.10% | 93.29% | **96.34%** |
| חלב | חֵלֶב | חָלָב | 78.24% | 75.29% | **85.77%** |
| חלה | חַלָּה | חָלָה | 86.80% | 94.32% | **96.80%** |
| חמור | חֲמוֹר | חָמוּר | 86.58% | 92.49% | **96.58%** |
| טבל | טָבַל | טֶבֶל | 85.92% | 89.08% | **94.97%** |
| נדר | נֶדֶר | נָדַר | 78.31% | 81.72% | **91.67%** |
| פה | פֶּה | פֹּה | 85.51% | 95.56% | **98.52%** |
| תנאי | תְּנַאי | תַּנָּאִי | 93.85% | 92.74% | **98.63%** |

Table 1: Evaluation of *Berel* on our Rabbinic Hebrew homograph challenge set, as compared with AlephBERT and Word2Vec. We report average F1 score for the two analyses of each homograph.

the mark together with the abbreviation as a single unit.

*Berel* was trained on a DGX Workstation, containing 4 A100 40GB GPUs, using the NVIDIA optimized extensions to the HuggingFace library. (Wolf et al., 2019)[3]. We trained the model only on training instances with fewer than 128 tokens. The model was trained with an initial learning rate of 6e-3 and a batch size of 8,192 training instances. Total training time was 9.5 days, during which the model completed 34,300 batches, a total of 14 epochs.

## 4 Rabbinic Hebrew Challenge Set

In order to evaluate the performance of *Berel* on Rabbinic texts, we created a homograph challenge set. The set focuses upon 12 homographs which recur frequently in Rabbinic texts, each of which admit to two common analyses. For each homograph, we select 300 random sentences from the Sefaria database containing the given homograph. Our expert annotators choose the correct analysis of the homograph in each case. The list of homographs is provided in Table 1.

## 5 Experiments

On the basis of this new challenge set, we contrast the performance of three approaches to homograph disambiguation:

- Word2Vec: We start with a pretrained set of Word2Vec embeddings for Rabbinic Hebrew. For each sentence, we encode of the context surrounding the homograph by running a Bi LSTM across the Word2Vec embeddings of the four words before and after the homograph. A classifier is trained to distinguish between the homograph analyses based on the Bi LSTM encoding. The Bi LSTM is trained together with the classifer itself, although the Word2Vec embeddings remain static as per their pretraining.

- AlephBERT: We run each sentence through AlephBERT and retrieve the embedding of the homograph token. We train a classifier based on those embeddings.

- *Berel*: Here too, we run each sentence through the model, retrieve the embedding of each homograph token, and we train a classifier based on those embeddings.

Within each of the three approaches, we train a 2-layer MLP for each one of the 12 homographs, and we evaluate performance via 10-fold cross-validation. For each homograph, we calculate precision and recall for each of the two analyses, and we compute the F1 score thereof. We then average the two F1 scores. Table 1 reports final scores for each of the 12 homographs and for each of the three evaluated methods. As can be seen, *Berel* consistently outperforms the other two approaches.

## 6 Resources for download

We provide the following resources for download:[4]

- Our pretrained Rabbinic Hebrew word2vec embeddings, trained on the Sefaria corpus (skipgram method, min freq = 7).

- Our Rabbinic Hebrew challenge set, containing 300 tagged sentences for each of 12 homographs.

- The *Berel* model itself, in pytorch format, along with our customized wordpiece tokenizer which includes handling for Rabbinic abbreviations and shortened words.

## 7 Conclusion

Heretofore, scholars and laymen who work with Rabbinic Hebrew have had to face a host of unsolved challenges. For instance, due to the frequent use of abbreviations in these texts, the abbreviations must be disambiguated before the text can be effectively processed. Additionally, because Rabbinic Hebrew texts are often printed without punctuation, the text must first be punctuated and divided into sentence-size chunks before it can be processed by other NLP algorithms. We believe that *Berel* will serve useful in solving these and other downstream tasks crucially needed for automated analysis of Rabbinic Hebrew.

---

[3] https://github.com/NVIDIA/DeepLearningExamples/tree/master/PyTorch/LanguageModeling/BERT

[4] https://bit.ly/3vzlvgG